\pdfoutput=1

\documentclass[11pt]{article}

\usepackage{emnlp2021}

\usepackage{times}
\usepackage{latexsym}
\usepackage{float}
\usepackage{graphicx}
\usepackage{booktabs}
\usepackage{amsmath,amsfonts,amssymb,amsthm}
\usepackage{bm}
\usepackage[ruled,noend]{algorithm2e}
\usepackage{enumitem}
\usepackage{setspace}

\usepackage[T1]{fontenc}

\newcommand{\stitle}[1]{\vspace{0.3em}\noindent{\bf #1}}
\SetKwInput{KwInput}{Input}
\SetKwInput{KwOutput}{Output}
\SetEndCharOfAlgoLine{}

\newcommand{\eg}{\emph{e.g.}\xspace} 

\usepackage{cleveref}
\crefformat{section}{\S#2#1#3}
\crefformat{subsection}{\S#2#1#3}
\crefformat{subsubsection}{\S#2#1#3}
\crefrangeformat{section}{\S\S#3#1#4 to~#5#2#6}
\crefmultiformat{section}{\S\S#2#1#3}{ and~#2#1#3}{, #2#1#3}{ and~#2#1#3}
\usepackage{refstyle}
\Crefformat{figure}{#2Fig.~#1#3}
\Crefmultiformat{figure}{Figs.~#2#1#3}{ and~#2#1#3}{, #2#1#3}{ and~#2#1#3}
\Crefformat{table}{#2Tab.~#1#3}
\Crefmultiformat{table}{Tabs.~#2#1#3}{ and~#2#1#3}{, #2#1#3}{ and~#2#1#3}
\Crefformat{appendix}{\S#2#1#3}
\crefformat{algorithm}{Alg.~#2#1#3}
\Crefformat{equation}{Eq.~#2#1#3}

\usepackage{microtype}

\title{Learning from Noisy Labels for Entity-Centric Information Extraction}

\author{Wenxuan Zhou \\
  University of Southern California \\
  \texttt{zhouwenx@usc.edu} \\\And
  Muhao Chen \\
  University of Southern California \\
  \texttt{muhaoche@usc.edu} \\}

\begin{document}
\maketitle
\begin{abstract}
Recent information extraction approaches have relied on training deep neural models.
However, such models can easily overfit noisy labels and suffer from performance degradation.
While it is very costly to filter noisy labels in large learning resources, recent studies show that such labels take more training steps to be memorized and are more frequently forgotten than clean labels, therefore are identifiable in training.
Motivated by such properties, we propose a simple co-regularization framework for entity-centric information extraction, which consists of several neural models with identical structures but different parameter initialization. 
These models are jointly optimized with the task-specific losses and are regularized to generate similar predictions based on an agreement loss, which prevents overfitting on noisy labels.
Extensive experiments on two widely used but noisy benchmarks for information extraction, TACRED and CoNLL03, demonstrate the effectiveness of our framework.
We release our code to the community for future research\footnote{Our code is publically available at \url{https://github.com/wzhouad/NLL-IE}}.
\end{abstract}

\section{Introduction}
Deep neural models have achieved significant success on various information extraction (IE) tasks.
However, when training labels contain noise, deep neural models can easily overfit the noisy labels, leading to severe performance degradation~\cite{Arpit2017ACL,Zhang2017UnderstandingDL}.
Unfortunately, 
labeling on large corpora, regardless of using human annotation~\cite{raykar2010learning} or automated heuristics~\cite{song2015spectral}, inevitably suffers from labeling errors.
This problem has even drastically affected widely used benchmarks, such as CoNLL03~\cite{Sang2003IntroductionTT} and TACRED~\cite{Zhang2017PositionawareAA}, 
where a notable portion of incorrect labels have been caused in annotation and largely hindered the performance of SOTA systems~\cite{reiss2020identifying,Alt2020TACREDRA}.
Hence, developing a robust learning method that better tolerates noisy supervision represents an urged challenge for emerging IE models.

So far, few research efforts have been made to developing noise-robust IE models,
and existing work mainly focuses on the weakly supervised or distantly supervised setting~\cite{Surdeanu2012MultiinstanceML,Ratner2016DataPC,huang2019self,mayhew2019named}.
\begin{figure}[t!]
    \centering
    \scalebox{0.21}{\includegraphics{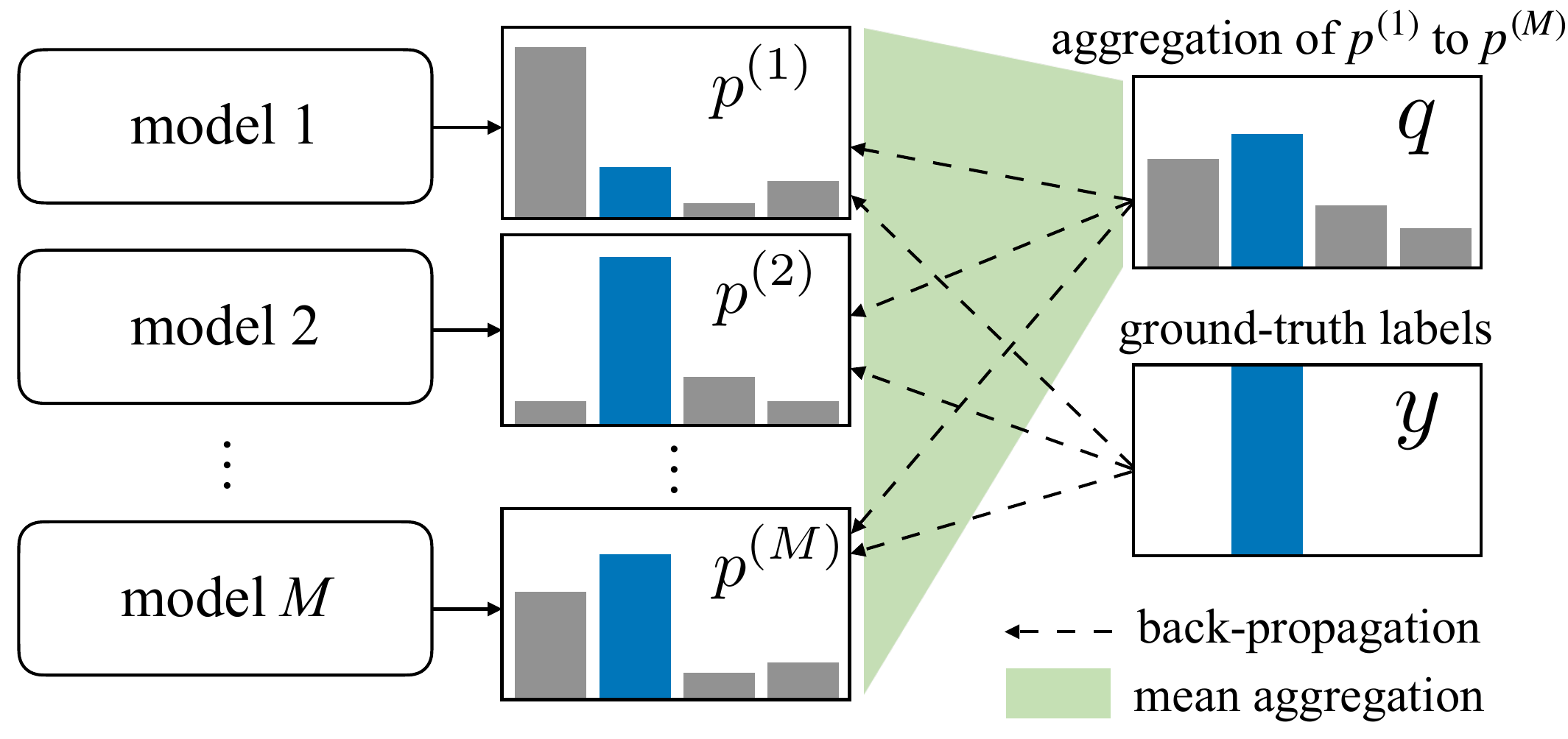}}
    \caption{Illustration of our co-regularization framework. The base models are jointly optimized with the task-specific loss from label $y$ and an agreement loss, which regularizes the models to generate similar predictions to the aggregated soft target probability $\bm{q}$.
}\label{fig:overview}
\end{figure}
Most of such methods typically depend on multi-instance learning that relies on bags of instances provided by distant supervision~\cite{Surdeanu2012MultiinstanceML,Zeng2015DistantSF, Ratner2016DataPC} or require an additional clean and sufficiently large reference dataset to develop a noise filtering model~\cite{Qin2018RobustDS}.
Accordingly, those methods may not be generally adapted to supervised training settings, where the aforementioned auxiliary learning resources are not always available.
Particularly, CrossWeigh~\cite{Wang2019CrossWeighTN} is a representative work that denoises a natural language dataset without using extra learning resources.
This method trains multiple independent models on different partitions of training data and downweighs instances on which the models disagree.
Though effective, a method of this kind requires training tens of redundant neural models, 
leading to excessive computational overhead for large models.
As far as we know, the problem of noisy labels in supervised learning for IE tasks has not been well investigated.

In this paper, we aim to develop a general denoising framework that can easily incorporate existing supervised learning models for entity-centric IE tasks.
Our method is motivated by studies~\cite{Arpit2017ACL,Toneva2019AnES} showing that noisy labels often have delayed learning curves, as incorrectly labeled instances are more likely to contradict the inductive bias captured by the model.
Hence, noisy label instances take a longer time to be picked up by neural models and are frequently forgotten in later epochs.
Therefore, predictions by more than one model tend to disagree on such instances.
Accordingly, we propose a simple yet effective co-regularization framework to handle noisy training labels, as illustrated in \Cref{fig:overview}.
Our framework consists of two or more neural classifiers with identical structures but different initialization.
In training, all classifiers are optimized on the training data with the task-specific loss and jointly regularized with regard to an agreement loss that is defined as the Kullback–Leibler (KL) divergence among predicted probability distributions.
Then for instances where a classifier's predictions disagree with labels, the agreement loss encourages the classifier to give similar predictions to the other classifier(s) instead of the actual (possibly noisy) labels.
In this way, the framework prevents the incorporated classifiers from overfitting noisy labels. 

We apply the framework to two important entity-centric IE tasks, named entity recognition (NER) and relation extraction (RE).
We conduct extensive experiments on two prevalent but noisy benchmarks, CoNLL03 for NER and TACRED for RE, and apply the proposed learning frameworks to train various models from prior studies for these two tasks. 
The results demonstrate the effectiveness of our method in noise-robust training, leading to promising and consistent performance improvement.
We present contributions as follows:
\begin{itemize}[leftmargin=1em]
    \setlength\itemsep{0em}
    \item We propose a general co-regularization framework that can effectively learn supervised IE models from noisy datasets without the need for any extra learning resources.
    \item We discuss in detail the different design strategies of the framework and the trade-off between efficiency and effectiveness.
    \item Extensive experiments on NER and RE demonstrate that our framework yields promising improvements on various SOTA models and outperforms existing denoising frameworks.
\end{itemize}

\section{Method}
In this paper, we focus on developing a noise-robust learning framework that improves supervised models for entity-centric IE tasks. 
In such tasks, (noisy) labels can be assigned to either 
individual tokens~(NER) or pairs of entities~(RE) in natural language text.
Specifically, $\mathcal{D}=\{(\bm{x}_i, y_i)\}_{i=1}^N$ is a noisily labeled dataset, 
where each data instance consists of a lexical sequence or a context $\bm{x}$, and a label $y$.
$y$ is annotated either on tokens of $\bm{x}$ for NER or on a pair of entity mentions in $\bm{x}$ for RE.
For some instances in $\mathcal{D}$, the labels are incorrect.
Our objective is to learn a noise-robust model $f$ with the presence of such noisily labeled instances from $\mathcal{D}$ without using external resources such as a clean development dataset~\cite{Qin2018RobustDS}.

\subsection{Learning Process}\label{sec:process}
Our framework is motivated by the delayed learning curve of a neural model on noisy data, compared with learning on clean data.
On noisy data, neural models tend to fit easy and clean instances that are more consistent with the well-represented patterns of data in early steps but need more steps to capture noise~\cite{Arpit2017ACL}.
Moreover, learned noisy examples tend to be frequently forgotten in later epochs~\cite{Toneva2019AnES} since they conflict with the general inductive bias represented by the clean data majority.
Therefore, model prediction is likely to be consistent with the clean labels while is often inconsistent or oscillates on noisy labels over different training epochs.
As a result, labels that are different from the model's predictions in the later epochs of training are likely to be noisy and should be down-weighted or rectified so as to reduce their impact on optimization.

The proposed framework incorporates several copies of a task-specific IE model with the same architecture but different (random) parameter initialization.
These IE models are jointly optimized on the noisy dataset based on their task-specific losses as well as on an agreement loss.
During training, the predicted probability distributions from models are aggregated as a soft target probability, which represents the models' estimations of the true label.
The agreement loss is responsible for encouraging these models to generate similar predictions to the soft target probability.
In this learning process,
models starting their training from varied initialization  
generate different decision boundaries.
By aggregating their predictions, the soft target probability can better separate noisy labels from clean labels that have not yet been learned.

The learning process of our framework is described in \Cref{algo::main}.
It consists of $M~(M\ge 2)$ copies of the task-specific model, denoted $\{f_k\}_{k=1}^M$, with different initialization.
Regarding initialization,
for models that are trained from scratch, all parameters are randomly initialized.
Otherwise, for those that are built upon pre-trained language models, only the parameters that are external to the language models~(\eg, those of a downstream softmax classifier) are randomly initialized, while the pre-trained parameters are the same.
Once initialized, our framework trains those models in two phases.
The first $\alpha\%$ training steps undergo a \emph{warm-up} phase, where $\alpha$ is a hyperparameter.
This phase seeks to help the model reach initial convergence on the task.
When a new batch comes in, we first calculate the task-specific training losses on $M$ models $\{\mathcal{L}_\text{sup}^{(k)}\}_{k=1}^M$ and average them as $\mathcal{L}_T$, then update model parameters w.r.t. $\mathcal{L}_T$.
After the warm-up phase, an agreement loss $\mathcal{L}_\text{agg}$ is further introduced to measure the distance from the predictions of $M$ models to the soft target probability $\bm{q}$. 
Parameters are accordingly updated based on the joint loss $\mathcal{L}$, encouraging the model to generate predictions that are consistent with both the training labels and the soft target probability.
The formalization of the loss function is described next (\Cref{sec:agreement}).
In the end, we can either use the model $f_1$ or select the best-performing model for inference.

\subsection{Co-regularization Objective}\label{sec:agreement}

{
\begin{algorithm}[!t]
    \caption{Learning Process}\label{algo::main}
    \setstretch{0.85}
    \normalsize
    \KwInput{Dataset $\mathcal{D}$, hyperparameters $T, \alpha, M, \gamma$.}
    \KwOutput{A trained model $f$.}
    Initialize $M$ neural models $\{f_k\}_{k=1}^M$. \\
    \For{$t=1...T$}{
    Sample a batch $\mathcal{B}$ from $\mathcal{D}$. \\
    Calculate task losses $\{\mathcal{L}_\text{sup}^{(k)}\}_{k=1}^M$. \\
    $\mathcal{L}_T = \frac{1}{M}\sum_{k=1}^M \mathcal{L}_\text{sup}^{(k)}$ \\
    \If(\tcp*[f]{Warmup}){$t < \alpha\% \times T$} 
    {
        Update model parameters w.r.t. $\mathcal{L}_T$. \\
    }
    \Else{
        Get the probability distribution of classes $\{\bm{p}\}_{k=1}^M$ with $M$ models. \\
        Calculate the soft target probability $\bm{q}$ by~\Cref{eq:q}. \\
        Calculate the agreement loss $\mathcal{L}_\text{agg}$ by~\Cref{eq:agreement1} and~\Cref{eq:agreement2}. \\
        $\mathcal{L}=\mathcal{L}_T + \gamma \cdot \mathcal{L}_\text{agg}$.\\
        Update model parameters w.r.t. $\mathcal{L}$. \\
    }
    }
    Return $f_1$ or the best-performing model.
\end{algorithm}
}

In our framework, the influence of noisy labels in training is decreased by optimizing the agreement loss.
Specifically, given a batch of data instances $\mathcal{B}=\{(\bm{x}_i, y_i)\}_{i=1}^N$, we first feed the instances to $M$ incorporated models to get their predictions $\left\{\{\bm{p}_i^{(k)}\}_{i=1}^N\right\}_{k=1}^M$ on $\mathcal{B}$, where $\bm{p}\in \mathbb{R}^C$ is the predicted probability distribution of $C$ classes.
Then we calculate the soft target probability $\bm{q}$ by averaging the predictions:
\begin{equation}\label{eq:q}
    \bm{q}_i = \frac{1}{M} \sum_{k=1}^M \bm{p}_i^{(k)},
\end{equation}
which represents the models' estimates of the true label.
Finally, we calculate the agreement loss $\mathcal{L}_\text{agg}$ as the average 
KL divergence from $\bm{q}$ to each $\bm{p}^{(k)}, k=1,..., M$:
\begin{align}
    &d(\bm{q}_i || \bm{p}_i^{(k)}) = \sum_{j=1}^C \bm{q}_{ij}\log\left(\frac{\bm{q}_{ij} + \epsilon}{\bm{p}_{ij} + \epsilon}\right), \label{eq:agreement1}\\
    &\mathcal{L}_\text{agg} = \frac{1}{MN} \sum_{i=1}^N \sum_{k=1}^M d(\bm{q}_i || \bm{p}_i^{(k)}), \label{eq:agreement2} 
\end{align}
where $\epsilon$ is a small positive number to avoid division by zero.
We can easily tell that the agreement loss encourages the models to get similar predictions based on the same input.
As the KL divergence is non-negative, the agreement loss is minimized only when $\bm{q}_i = \bm{p}_i^{(k)}$ for $k=1,...,M$, which implies that all $\bm{p}_i^{(k)}$ should be equal because we use the average probability for $\bm{q}$.
We may also use other aggregates for $\bm{q}$ as long as they satisfy that $\bm{q}_i = \bm{p}_i^{(k)}, k=1,...,M$ when all $\bm{p}_i^{(k)}$ are equal so as to maintain such property of the agreement loss.
We consider the following alternatives for $\bm{q}$:
\begin{itemize}[leftmargin=1em]
    \setlength\itemsep{0em}
    \item \textbf{Average logits.} Given the logits $\{\bm{l}_i^{(k)}\}_{k=1}^M$ of predicted probabilities on the $M$ models, we first average the logits $\bm{l}_i = \frac{1}{M} \sum_{k=1}^M \bm{l}_i^{(k)}$ and then feed $\bm{l}_i$ to a softmax function to get the soft target probability $\bm{q}$.

    \item \textbf{Max-loss probability.} A noise-robust model will disagree on noisy labels and produce large training losses.
    Therefore, for each instance $i$ in the batch, we assume the prediction $p_i^*$ that has the largest task-specific loss among the $M$ models to be more reliable and use it as the soft target probability for instance $i$.
\end{itemize}
In experiments, we observe that all aggregate functions generally achieve similar performance.
We present the results of different $\bm{q}$ in \Cref{sec:ablation}.

\subsection{Joint Training}
The main learning objective of our framework is then to optimize the joint loss $\mathcal{L}=\mathcal{L}_T+\gamma\mathcal{L}_\text{agg}$, where $\gamma$ is a positive hyperparameter and $\mathcal{L}_T$ is the average of task-specific classification losses $\left\{\mathcal{L}_\text{sup}^{(k)}\right\}_{k=1}^M$. 
For classification problems such as NER and RE, the task-specific loss is defined as the following cross-entropy loss, where $\bm{I}$ denotes an indicator function:
\begin{align}
    \mathcal{L}_\text{sup} = -\frac{1}{N}\sum_{i=1}^N\sum_{j=1}^C \bm{I}\left[y_i = j\right] \log \bm{p}_{ij}.
\end{align}
$N$ thereof is the number of tokens for NER and the number of sentences for RE.

The joint training can be interpreted as a ``soft-pruning'' scheme.
For clean labels where the models' predictions are usually close to the labels, the agreement loss and its gradient are both small, so they have a small impact on training.
While for noisy labels where the model predictions disagree with the training labels, the agreement loss incurs a large magnitude of gradients in training, which prevents the model from overfitting the noisy labels.

Besides co-regularization, denoising may also be attempted by ``hard-pruning'' the noisy labels.
Small-loss selection~\cite{Jiang2018MentorNetLD,Han2018CoteachingRT} assumes that instances with large task-specific loss are noisy and excludes them from training.
However, some clean label instances, especially those from long-tail classes, can also have large task-specific losses and will be incorrectly pruned.
While for the frequent classes, some noisy instances can have smaller task-specific losses and fail to be identified.
Such errors can accumulate during training and may hinder model performance.
In our framework, as we use the agreement loss instead of hard pruning, such errors will not be easily propagated~(see \Cref{sec:ablation}).

\section{Tasks}


We evaluate our framework on two fundamental entity-centric IE tasks, namely RE and NER.
Our framework can incorporate any kind of neural model that is dedicated to either task.
Particularly, in this paper, we adopt off-the-shelf SOTA models that are mainly based on Transformers.
This section introduces the two attempted tasks and the design of task-specific models.

\stitle{Relation extraction}.
RE aims at identifying the relations between a pair of entities in a piece of text from the given vocabulary of relations.
Specifically, given a sentence $\bm{x}$ and two entities $e_s$ and $e_o$, identified as the subject and object entities respectively, the goal is to predict the relation between $e_s$ and $e_o$.
Following \citet{Shi2019SimpleBM}, we formulate this task as a sentence classification problem.
Accordingly, we first apply the entity masking technique~\cite{Zhang2017PositionawareAA} to the input sentence and replace the subject and object entities with their named entity types.
For example, a short sentence ``\textit{Bill Gates founded Microsoft}'' will become ``\textsc{[subject-person]} \textit{founded} \textsc{[object-organization]}'' after entity masking.
We then feed the sentence to the pre-trained language model and use a softmax classifier on the representation of the \textsc{[cls]} token to predict the relation.

\stitle{Named entity recognition}. NER seeks to locate and classify named entities in text into pre-defined categories.
Following \citet{Devlin2019BERTPO}, we formulate the task as a token classification problem.
In detail, 
a Transformer-based language model first tokenizes an input sentence into a sub-token sequence.
To classify each token, the representation of its first sub-token is sent into a softmax classifier.
We use the BIO tagging scheme~\cite{Ramshaw1995TextCU} and output the tag with the maximum likelihood as the predicted label.

\section{Experiment}


In this section, we evaluate the proposed learning framework based on two (noisy) benchmark datasets for the two entity-centric IE tasks (\Cref{ssec:dataset}-\Cref{ssec:main_results}).
In addition, a noise filtering analysis is presented to show how our framework prevents an incorporated neural model from overfitting noisy training data (\Cref{ssec:noise_filtering}), along with a detailed ablation study about configurations with varied model copies, alternative noise filtering strategies, target functions, and different noise rates (\Cref{sec:ablation}).

\subsection{Datasets}\label{ssec:dataset}

The experiments are conducted on TACRED~\cite{Zhang2017PositionawareAA} and CoNLL03~\cite{Sang2003IntroductionTT}. TACRED is a crowdsourced dataset for relation extraction.
A recent study by \citet{Alt2020TACREDRA} found a large portion of examples to be mislabeled and rectified some incorrect labels in the development and test sets. CoNLL03 is a human-annotated dataset for NER.
Another study by \citet{Wang2019CrossWeighTN} found that in 5.38\% of sentences in CoNLL03, at least one token is mislabeled. 
Accordingly, \citeauthor{Wang2019CrossWeighTN} also relabeled the test set\footnote{Note that neither of these two datasets comes with a relabeled (clean) training set. All models are still trained on the original noisy training set.}.
We summarize the statistics of both datasets in \Cref{tab:data_statistics}.
For all compared methods, we report the results on both the original and relabeled evaluation sets.

\subsection{Base Models}\label{sup:comp}
We evaluate our framework by incorporating the following SOTA models:
\begin{itemize}[leftmargin=1em]
    \setlength\itemsep{0em}
    \item \textbf{C-GCN}~\cite{Zhang2018GraphCO} is a graph-based model for RE.
    It prunes the dependency graph and applies graph convolutional networks to get the representation of entities.
    \item \textbf{BERT}~\cite{Devlin2019BERTPO} is a Transformer-based language model that is pre-trained from large-scale text corpora.
    Both \texttt{Base} and \texttt{Large} versions of the model are considered in our experiments.
    \item \textbf{LUKE}~\cite{Yamada2020LUKEDC} is a Transformer-based language model that is pre-trained on both large-scale text corpora and knowledge graphs.
    It achieves SOTA performance on various entity-related tasks, including RE and NER.
\end{itemize}
We report the performance of the base models trained with and without our co-regularization framework.
We also compare our framework to CrossWeigh~\cite{Wang2019CrossWeighTN}, which is another noisy-label learning framework.
Specifically, CrossWeigh partitions the training set into equal-sized chunks, reserves each chunk, and then trains several models on the rest ones.
After training, the models predict on the reserved chunk, and instances on which the models disagree are down-weighted.
In the end, the chunks are combined and used to train a new model for inference.
Learning by CrossWeigh is dependant on a high computation cost.
\citet{Wang2019CrossWeighTN} split the CoNLL03 dataset into 10 chunks and train 3 models on each partition, resulting in a total of number 30 models.
In this paper, we follow their settings and train 30 models on both TACRED and CoNLL03.

\begin{table}[t]
    \centering
    \scalebox{0.74}{
    \begin{tabular}{p{1.6cm}ccccc}
    \toprule
    \textbf{Dataset}& \# train& \# dev& \# test& \# classes& \% noise \\
    \midrule
    TACRED& 68124& 22631& 15509& 42& 6.62\\
    CoNLL03& 14041& 3250& 3453& 9& 5.38 \\
    \bottomrule
    \end{tabular}}
    \caption{Data statistics of TACRED and CoNLL03.}
    \label{tab:data_statistics}
\end{table}

\subsection{Model Configurations}\label{sup:conf}
For base models C-GCN~\cite{Zhang2018GraphCO} and LUKE~\cite{Yamada2020LUKEDC}, we rerun the officially released implementations using the recommended hyperparameters in the original papers.
We implement BERT$_\text{BASE}$ and BERT$_\text{LARGE}$ based on
Huggingface's Transformers \cite{wolf2020transformers}.
For CrossWeigh~\cite{Wang2019CrossWeighTN}, we re-implement this framework using those compared base models.
All models are optimized with Adam~\cite{Kingma2015AdamAM} using a learning rate of $6\mathrm{e}{-5}$ for \mbox{TACRED} and that of  $1\mathrm{e}{-5}$ for \mbox{CoNLL03}, with a linear learning rate decay to 0.
The batch size is fixed as 64 for all models.
We finetune the TACRED model for 5 epochs and the CoNLL03 model for 50 epochs.
The best model checkpoint is chosen based on the $F_1$ score on the development set.
We tune $\gamma$ from $\{1.0, 2.0, 5.0, 10.0, 20.0\}$, and tune $\alpha$ from $\{10, 30, 50, 70, 90\}$.
We report the median of $F_1$ of 5 runs using different random seeds.

For efficiency, we use the simplest setup of our framework with two model copies ($M=2$) in the main experiments (\Cref{ssec:main_results}).
$\bm{q}$ is set as the average probability in the main experiment.
Performance with more model copies and alternative aggregates is later studied in \Cref{sec:ablation}.

\begin{table}[!t]
\centering
\scalebox{0.73}{
\setlength{\tabcolsep}{2pt}
    \begin{tabular}{lcccc}
         \toprule
         \textbf{Model} & \multicolumn{2}{c}{\textbf{Original}} & \multicolumn{2}{c}{\textbf{Relabeled}} \\
          &Dev $F_1$ & Test $F_1$ & Dev $F_1$& Test $F_1$ \\
         \midrule
         C-GCN $\clubsuit$~\cite{Zhang2018GraphCO}& 67.2& 66.7& 74.9& 74.6 \\
         C-GCN-CrossWeigh& 67.8& 67.4& 75.6& 75.7 \\
         C-GCN-CR& 67.7& 67.2& 75.6&  75.4\\
        \midrule
         BERT$_\text{BASE}$~\cite{Devlin2019BERTPO}& 69.1& 68.9& 76.4& 76.9 \\
         BERT$_\text{BASE}$-CrossWeigh& 71.3& 70.8& 79.2& 79.1 \\
         BERT$_\text{BASE}$-CR& 71.5& 71.1& 79.9& 80.0\\
         \midrule
         BERT$_\text{LARGE}$~\cite{Devlin2019BERTPO}& 70.9& 70.2& 78.3& 77.9 \\
         BERT$_\text{LARGE}$-CrossWeigh& 72.1& 71.9& 79.5& 79.8 \\
         BERT$_\text{LARGE}$-CR& 73.1& 73.0& 81.3& 82.0 \\
         \midrule
         LUKE $\clubsuit$~\cite{Yamada2020LUKEDC}& 71.1& 70.9& 80.1& 80.6 \\
         LUKE-CrossWeigh& 71.0& 71.6& 80.4& 81.6 \\
         LUKE-CR& 71.8& 72.4& 81.9& 83.1 \\
         \bottomrule
    \end{tabular}
    }
    \caption{$\bm{F}_1$ score (\%) on the dev and test set of TACRED. $\clubsuit$ marks results obtained from the originally released implementation. We report the median of $F_1$ on 5 runs of training using different random seeds. For fair comparison, the CR results are reported based on the predictions from model $f_1$ in our framework.
}\label{tab::re}
\end{table}

\subsection{Main Results}\label{ssec:main_results}
The experiment results on TACRED and CoNLL03 are reported in \Cref{tab::re} and \Cref{tab::ner} respectively, where methods incorporated in our learning framework are marked with ``CR''.
As stated, the results are reported under the setup where $M=2$.
For a fair comparison, the results are reported based on the predictions from model $f_1$ in the framework.
On TACRED, our framework leads to an absolute improvement of $2.5-4.1\%$ in
$F_1$ on the relabeled test set for Transformer-based models, and a relatively smaller gain (0.8\% in $F_1$) for C-GCN.
In particular, our framework enhances the SOTA method LUKE by 2.5\% in $F_1$, leading to a very promising $F_1$ score of 83.1\%.
On CoNLL03, where the noise rate is smaller than TACRED, our framework leads to a performance gain of $0.28-0.82\%$ in $F_1$ on the relabeled test set.
On both IE tasks, our framework also leads to a consistent improvement on the original test set.
Compared to CrossWeigh, except for C-GCN where the results are similar, our framework consistently outperforms it by $0.9-2.2\%$ on TACRED and by $0.13-0.45\%$ on CoNLL03.
Moreover, as our framework requires training $M$ models concurrently while CrossWeigh requires training redundant models~(30 in experiments), the computation cost of our co-regularization framework is much lower than CrossWeigh.
In general, the results here show the effectiveness and practicality of the proposed framework.

\subsection{Noise Filtering Analysis}\label{ssec:noise_filtering}
The main experiments show that our framework can improve the overall performance of models trained with noisy labels.
In this section, we further demonstrate how our framework prevents overfitting on noisy labels.
To do so, we extract the 2,526 noisy instances from the development and test sets of TACRED where the relabeling by \citet{Alt2020TACREDRA} disagrees with the original labels.
Accordingly, we obtain a \emph{noisy set} containing those examples with original labels and a \emph{clean set} with rectified labels. 
We train a relation classifier on the union of the training set and the noisy set and then evaluate the model on the clean set.
In this case, worse performance on the clean set indicates more severe overfitting on noisy labels.

\begin{figure}[t!]
    \centering
    \scalebox{0.55}{\includegraphics{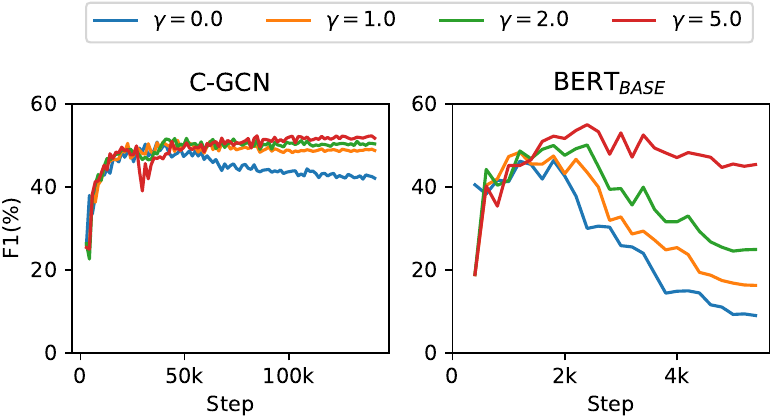}}
    \caption{$\bm{F}_1$ score (\%) on the clean set of TACRED. Classifiers trained with our framework are more noise-robust compared to baselines ($\gamma=0$).}\label{fig:noise}
\end{figure}

\Cref{fig:noise} shows the results by C-GCN-CR and BERT$_\text{BASE}$-CR on the clean set, where we observe that:
(1) Compared to the original base models~($\gamma=0.0$), those trained with our framework achieves higher $F_1$ scores, indicating improved robustness against the label noise;
(2) Comparing different base models, the large classifier BERT$_\text{BASE}$ is typically less noise-robust than a smaller model like C-GCN, 
which explains why the performance gain from our framework is more notable on BERT$_\text{BASE}$;
(3) For both models, the $F_1$ score first increases then decreases, consistent with the delayed learning curves that the neural models have on noisy instances~\cite{Arpit2017ACL}.

\begin{table}[!t]
\centering
\scalebox{0.73}{
\setlength{\tabcolsep}{2pt}
    \begin{tabular}{lccc}
         \toprule
         \textbf{Model} & \multicolumn{2}{c}{\textbf{Original}} & \multicolumn{1}{c}{\textbf{Relabeled}} \\
          & Dev $F_1$ &Test $F_1$ & Test $F_1$ \\
         \midrule
         BERT$_\text{BASE}$~\cite{Devlin2019BERTPO}&95.58& 91.96& 92.91 \\
         BERT$_\text{BASE}$-CrossWeigh& 95.65& 92.15& 93.03\\
         BERT$_\text{BASE}$-CR& 95.87& 92.53& 93.48 \\
         \midrule
         BERT$_\text{LARGE}$~\cite{Devlin2019BERTPO}& 96.16& 92.24& 93.22 \\
         BERT$_\text{LARGE}$-CrossWeigh& 96.32& 92.49& 93.61 \\
         BERT$_\text{LARGE}$-CR& 96.59& 92.82& 94.04 \\
         \midrule
         LUKE $\clubsuit$~\cite{Yamada2020LUKEDC}& 97.03& 93.91& 95.60 \\
         LUKE--CrossWeigh& 97.09& 93.98& 95.75\\
         LUKE-CR& 97.21& 94.22& 95.88 \\
         \bottomrule
    \end{tabular}}
    \caption{$\bm{F}_1$ score (\%) on the dev and test set of CoNLL03. $\clubsuit$ marks results obtained using the originally released code.}\label{tab::ner}
\end{table}

\subsection{Ablation Study}\label{sec:ablation}

\stitle{Using extra model copies.} The main results show that using two copies of a model in the co-regularization framework has already improved the performance by a remarkable margin.
Intuitively, more models may generate higher-quality soft target probabilities and thus further improve the performance.
We further show the performance on TACRED by incorporating more model copies.
We report the relabeled test $F_1$ on TACRED in~\Cref{tab:more_models}.
We observe that increasing the number of copies does not necessarily lead to a notable increase in performance.
On BERT$_\text{LARGE}$, increasing the number of model copies from 2 to 4 gradually improves the performance from $82.0\%$ to $82.7\%$.
While on BERT$_\text{BASE}$, increasing the number of model copies does not improve the performance.
We notice that the increased number of copies leads to a significant increase in the agreement loss for BERT$_\text{BASE}$, indicating that the copies of BERT$_\text{BASE}$ fail to reach a consensus based on the same input.
This may be due to the relatively small model capacity of BERT$_\text{BASE}$.
Overall, this study shows that the optimal $M$ is dependent on the models and needs to be tuned on the specific task.
Note that as the models can be trained in parallel, increasing the number of models does not necessarily increase the training time, though being at the cost of more computational resources.

\begin{table}[t]
    \centering
    \scalebox{0.78}{
    \begin{tabular}{p{3cm}ccc}
    \toprule
     \textbf{\# Models} & 2& 3& 4 \\
     \midrule
     BERT$_\text{BASE}$-CR& 80.0& 79.5& 79.8 \\
     BERT$_\text{LARGE}$-CR& 82.0& 82.4& 82.7\\
     \bottomrule
    \end{tabular}}
    \caption{$\bm{F_1}$ score (\%) of using different number of models on the relabeled test set of TACRED.}
    \label{tab:more_models}
\end{table}

\begin{table}[t]
    \centering
    \scalebox{0.74}{
    \begin{tabular}{lcc}
    \toprule
     \textbf{Model} & \multicolumn{1}{c}{\textbf{Original}} & \multicolumn{1}{c}{\textbf{Relabeled}} \\
     & Test $F_1$& Test $F_1$ \\
     \midrule
     BERT$_\text{BASE}$& 68.9& 76.9 \\
     BERT$_\text{BASE}$ + Small-loss selection& 68.7& 76.6 \\
     BERT$_\text{BASE}$ + Relabeling& 69.0& 77.7 \\
    \bottomrule
    \end{tabular}}
    \caption{$\bm{F_1}$ score (\%) of alternative noise filtering strategies on the test set of TACRED. The best results are achieved when $\delta=2\%$ for both methods.}
    \label{tab:ablation}
\end{table}

\begin{table}[t]
    \centering
    \scalebox{0.75}{
    \begin{tabular}{p{3cm}ccc}
    \toprule
     \textbf{Functions} & Avg prob& Avg logit& Max-loss prob \\
     \midrule
     BERT$_\text{BASE}$-CR& 80.0& 79.9& 79.4\\
     BERT$_\text{LARGE}$-CR& 82.0& 81.6& 82.2\\
     \bottomrule
    \end{tabular}}
    \caption{$\bm{F_1}$ score (\%) of different functions for $\bm{q}$ on the relabeled test set of TACRED.}
    \label{tab:other_functions}
\end{table}

\begin{table}[t]
    \centering
    {\small
    \setlength{\tabcolsep}{3pt}
    \begin{tabular}{lccccc}
    \toprule
     \textbf{Flipped labels ($\%$)} & $10$& $30$& $50$& $70$& $90$ \\
     \midrule
     BERT$_\text{BASE}$& 74.2& 70.8& 62.9& 48.6& 0\\
     BERT$_\text{BASE}$-CrossWeigh& 77.3& 75.6& 71.6& 61.3& 25.1\\
     BERT$_\text{BASE}$-CR& 79.3& 78.3&
     73.2& 63.5& 34.1\\
     \midrule
     BERT$_\text{BASE}$ w/o flipped labels& 76.5& 74.9& 72.9& 70.8& 57.4\\
    \bottomrule
    \end{tabular}
    }
    \caption{$\bm{F_1}$ score (\%) under different noise rates on the relabeled set of TACRED.}
    \label{tab:noise_rates}
\end{table}

\stitle{Alternative strategies for noise filtering.}
Besides co-regularization, we also experiment with other noise-filtering strategies.
Small-loss selection~\cite{Jiang2018MentorNetLD,Han2018CoteachingRT,lee2019robust} prunes the instances with the largest training losses in the training batches.
This method is motivated by the fact that the noisy instances take a longer time to be memorized and usually cause a large training loss.
We further try another strategy named relabeling.
Instead of pruning the large-loss training instances, we relabel them with the most likely labels from model predictions.

We evaluate the two noise filtering strategies on TACRED using BERT$_\text{BASE}$ as the base model.
For both strategies, we prune/relabel $\delta_t=\delta \cdot \frac{t}{T}$ percent of examples with the largest training loss in each training batch following~\citet{Han2018CoteachingRT}, where $t$ is the current number of training steps, $T$ is the total number of training steps, and $\delta$ is the maximum pruning/relabeling rate.
These hyperparameters are tuned on the development set.
The training loss is defined as the average task-specific loss of the $M$ models, where we set $M=2$ in consistent with the main experiments (\Cref{ssec:main_results}).
We try $\delta$ from $\{2\%, 5\%, 8\%\}$ and report the best results.

Results are shown in Table~\ref{tab:ablation}.
We find that $\delta=2\%$ achieves the best performance for both strategies.
The small-loss selection strategy underperforms the base model without noise filtering.
Relabeling outperforms the base model slightly, but the improvements are lesser than the proposed co-regularization method.
We observe that these two strategies do not work well on imbalanced datasets, mostly pruning or relabeling training examples from long-tail classes.
Specifically, on the TACRED dataset, where the \textsc{na} class accounts for $~80\%$ of the total labels, only $20\%$ pruned labels are from \textsc{na} while the remaining $80\%$ are from other classes.
It is because that the model's predictions will be biased towards the frequent classes on imbalanced datasets, therefore leading to the large training loss on long-tail instances.
Once pruned or relabeled, such long-tail instances are excluded from training, causing further error propagation that can lead to more biased predictions.
Our framework, on the contrary, adopts an agreement loss instead of hard pruning or relabeling, which reduces such error propagation.

\stitle{Alternative aggregates for $\bm{q}$.}
Besides the average probability, we evaluate two other aggregates for $\bm{q}$, i.e. the average logits and the max-loss probability (\Cref{sec:agreement}).
This experiment is conducted with $M=2$. 
$F_1$ results on the relabeled TACRED test set (\Cref{tab:other_functions}) suggest that different aggregates generally achieve comparable performance,
with a marginal difference of up to $0.6\%$ in $F_1$.
Therefore, the default setup is suggested to be the average probability, which is easier to implement.

\stitle{Performance under different noise rates.}
We further evaluate our framework on training data of different noise rates.
To do so, we create noisy training data by randomly flipping $10\%$, $30\%$, $50\%$, $70\%$, or $90\%$ labels in the training set of TACRED.
Then we use those synthetic noisy training sets to train RE models and evaluate them on the relabeled test set of TACRED.
We use BERT$_\text{BASE}$ as the base model and report the median $F_1$ score of 5 trials.
Results are given in~\Cref{tab:noise_rates},
which show that our co-regularization framework consistently outperforms both the base model and CrossWeigh under different noise rates. 
The gain generally becomes larger as the noise rate increases.
In comparison to BERT$_\text{BASE}$ trained on the training sets where all flipped labels are removed, our framework, even trained on synthetic noise, achieves comparable or better results when the noise rates are below $50\%$.

\section{Related Work}

We discuss two lines of related work.
Each has a large body of work which we can only provide as a highly selected summary.

\stitle{Distant supervision}. Distant supervision~\cite{Mintz2009DistantSF} generates noisy training data with heuristics to align unlabeled data with labels, whereas much effort has been devoted to reducing labeling noise.
Multi-instance learning~\cite{Zeng2015DistantSF,Lin2016NeuralRE,Ji2017DistantSF,Liu2017ASM} creates bags of noisily labeled instances and assumes at least one instance in each bag is correct, then it uses heuristics or auxiliary classifiers to select the correct labels.
However, such instance bags may not exist in a general supervised setting.
Reinforcement learning~\cite{Qin2018RobustDS,Yang2019ExploitingND,Wang2020FindingII} and curricular learning~\cite{Jiang2018MentorNetLD,huang2019self} methods use a clean validation set to obtain an auxiliary model for noise filtering, while constructing a perfectly labeled validation set is expensive.
Our framework can learn noise-robust IE models without extra learning resources and can be easily incorporated into existing supervised IE models.

\stitle{Supervised learning with noisy labels}.
A deep neural network can memorize noisy 
labels, and its 
generalizability will severely degrade when trained 
with noisy labels~\cite{Zhang2017UnderstandingDL}.
In 
computer vision, 
much investigation has been 
conducted for supervised image classification with noise,
producing techniques such as robust loss functions~\cite{Zhang2018GeneralizedCE,Wang2019SymmetricCE}, noise filtering layers~\cite{Sukhbaatar2014TrainingCN,Goldberger2017TrainingDN}, label re-weighting~\cite{wang2017multiclass,Chang2017ActiveBT}, robust regularization~\cite{krogh1992simple,srivastava2014dropout,muller2019does}, and sample selection~\cite{Malach2017DecouplingT,Jiang2018MentorNetLD,Han2018CoteachingRT,Yu2019HowDD,Wei2020CombatingNL}.
The robust loss functions and noise filtering layers require modifying model structures and may not be easily adapted to IE models.
The sample selection methods assume the data instances with large training losses to be noisy and exclude them from training.
However, some clean instances, especially those from long-tail classes, can also have a large training loss and be wrongly pruned, leading to propagated errors.

In NLP, few efforts have focused on learning with denoising.
CrossWeigh~\cite{Wang2019CrossWeighTN}, one label re-weighting method, partitions the training data into multiple folds and trains multiple models on each fold.
Instances on which models disagree are regarded as noisy and down-weighted in training.
However, this method requires training many models and is computationally expensive.
Our framework only requires training several models concurrently, which is more computationally efficient and achieves better performance.
NetAb~\cite{Wang2019LearningWN} assumes that noisy labels are created from randomly flipping clean labels and uses a CNN to model the noise transition matrix~\cite{Wang2018MulticlassLW}.
However, this assumption does not hold for real datasets, where the noise rate vary among data instances~\cite{Cheng2017LearningWB}.

\section{Conclusion}
This paper presents a co-regularization framework for learning supervised IE models from noisy data.
This framework consists of two or more identically structured models with different initialization, which are encouraged to give similar predictions on the same inputs by optimizing an agreement loss.
On noisy examples where model predictions usually differ from the labels, the agreement loss prevents the model from overfitting noisy labels.
Experiments on NER and RE benchmarks show that our framework yields promising improvements on various IE models.
For future work, we plan to extend the use of the proposed framework to other tasks such as event-centric IE \cite{chen2021event} and co-reference resolution \cite{peng2015solving}.

\section*{Ethical Consideration}
This work does not present any direct societal consequence. The proposed work seeks to develop a general learning framework that learning more robust neural models for entity-centric information extraction under noisy label settings. We believe this leads to intellectual merits that benefit the information extraction community where learning resources may often suffer from noisy labeling issues. And it potentially has broad impacts since the tackled issues also widely exist in tasks of other areas.

\section*{Acknowledgment}

We appreciate the anonymous reviewers for their insightful comments and suggestions. 
Also, we would like to thank Fangyu Liu from the Language Technology Lab at the University of Cambridge for the discussion and inputs when writing this paper.

This material is supported in part by the DARPA MCS program under Contract No. N660011924033 with the United States Office Of Naval Research, and by the National Science Foundation of United States Grant IIS 2105329.

\bibliography{anthology,custom}

\begin{thebibliography}{49}
\expandafter\ifx\csname natexlab\endcsname\relax\def\natexlab#1{#1}\fi

\bibitem[{Alt et~al.(2020)Alt, Gabryszak, and Hennig}]{Alt2020TACREDRA}
Christoph Alt, Aleksandra Gabryszak, and Leonhard Hennig. 2020.
\newblock {TACRED} revisited: A thorough evaluation of the {TACRED} relation
  extraction task.
\newblock In \emph{Proceedings of the 58th Annual Meeting of the Association
  for Computational Linguistics}, pages 1558--1569, Online. Association for
  Computational Linguistics.

\bibitem[{Arpit et~al.(2017)Arpit, Jastrzebski, Ballas, Krueger, Bengio,
  Kanwal, Maharaj, Fischer, Courville, Bengio, and
  Lacoste{-}Julien}]{Arpit2017ACL}
Devansh Arpit, Stanislaw Jastrzebski, Nicolas Ballas, David Krueger, Emmanuel
  Bengio, Maxinder~S. Kanwal, Tegan Maharaj, Asja Fischer, Aaron~C. Courville,
  Yoshua Bengio, and Simon Lacoste{-}Julien. 2017.
\newblock A closer look at memorization in deep networks.
\newblock In \emph{Proceedings of the 34th International Conference on Machine
  Learning, {ICML} 2017, Sydney, NSW, Australia, 6-11 August 2017}, volume~70
  of \emph{Proceedings of Machine Learning Research}, pages 233--242. {PMLR}.

\bibitem[{Chang et~al.(2017)Chang, Learned-Miller, and
  McCallum}]{Chang2017ActiveBT}
Haw-Shiuan Chang, E.~Learned-Miller, and A.~McCallum. 2017.
\newblock Active bias: Training more accurate neural networks by emphasizing
  high variance samples.
\newblock In \emph{NeurIPS}.

\bibitem[{Chen et~al.(2021)Chen, Zhang, Ning, Li, Ji, McKeown, and
  Roth}]{chen2021event}
Muhao Chen, Hongming Zhang, Qiang Ning, Manling Li, Heng Ji, Kathleen McKeown,
  and Dan Roth. 2021.
\newblock Event-centric natural language processing.
\newblock In \emph{Proceedings of the 59th Annual Meeting of the Association
  for Computational Linguistics}.

\bibitem[{Cheng et~al.(2020)Cheng, Liu, Ramamohanarao, and
  Tao}]{Cheng2017LearningWB}
Jiacheng Cheng, Tongliang Liu, Kotagiri Ramamohanarao, and Dacheng Tao. 2020.
\newblock Learning with bounded instance and label-dependent label noise.
\newblock In \emph{Proceedings of the 37th International Conference on Machine
  Learning, {ICML} 2020, 13-18 July 2020, Virtual Event}, volume 119 of
  \emph{Proceedings of Machine Learning Research}, pages 1789--1799. {PMLR}.

\bibitem[{Devlin et~al.(2019)Devlin, Chang, Lee, and
  Toutanova}]{Devlin2019BERTPO}
Jacob Devlin, Ming-Wei Chang, Kenton Lee, and Kristina Toutanova. 2019.
\newblock {BERT}: Pre-training of deep bidirectional transformers for language
  understanding.
\newblock In \emph{Proceedings of the 2019 Conference of the North {A}merican
  Chapter of the Association for Computational Linguistics: Human Language
  Technologies, Volume 1 (Long and Short Papers)}, pages 4171--4186,
  Minneapolis, Minnesota. Association for Computational Linguistics.

\bibitem[{Goldberger and Ben{-}Reuven(2017)}]{Goldberger2017TrainingDN}
Jacob Goldberger and Ehud Ben{-}Reuven. 2017.
\newblock Training deep neural-networks using a noise adaptation layer.
\newblock In \emph{5th International Conference on Learning Representations,
  {ICLR} 2017, Toulon, France, April 24-26, 2017, Conference Track
  Proceedings}. OpenReview.net.

\bibitem[{Han et~al.(2018)Han, Yao, Yu, Niu, Xu, Hu, Tsang, and
  Sugiyama}]{Han2018CoteachingRT}
Bo~Han, Quanming Yao, Xingrui Yu, Gang Niu, Miao Xu, Weihua Hu, Ivor~W. Tsang,
  and Masashi Sugiyama. 2018.
\newblock Co-teaching: Robust training of deep neural networks with extremely
  noisy labels.
\newblock In \emph{Advances in Neural Information Processing Systems 31: Annual
  Conference on Neural Information Processing Systems 2018, NeurIPS 2018,
  December 3-8, 2018, Montr{\'{e}}al, Canada}, pages 8536--8546.

\bibitem[{Huang and Du(2019)}]{huang2019self}
Yuyun Huang and Jinhua Du. 2019.
\newblock Self-attention enhanced {CNN}s and collaborative curriculum learning
  for distantly supervised relation extraction.
\newblock In \emph{Proceedings of the 2019 Conference on Empirical Methods in
  Natural Language Processing and the 9th International Joint Conference on
  Natural Language Processing (EMNLP-IJCNLP)}, pages 389--398, Hong Kong,
  China. Association for Computational Linguistics.

\bibitem[{Ji et~al.(2017)Ji, Liu, He, and Zhao}]{Ji2017DistantSF}
Guoliang Ji, Kang Liu, Shizhu He, and Jun Zhao. 2017.
\newblock Distant supervision for relation extraction with sentence-level
  attention and entity descriptions.
\newblock In \emph{Proceedings of the Thirty-First {AAAI} Conference on
  Artificial Intelligence, February 4-9, 2017, San Francisco, California,
  {USA}}, pages 3060--3066. {AAAI} Press.

\bibitem[{Jiang et~al.(2018)Jiang, Zhou, Leung, Li, and
  Fei{-}Fei}]{Jiang2018MentorNetLD}
Lu~Jiang, Zhengyuan Zhou, Thomas Leung, Li{-}Jia Li, and Li~Fei{-}Fei. 2018.
\newblock Mentornet: Learning data-driven curriculum for very deep neural
  networks on corrupted labels.
\newblock In \emph{Proceedings of the 35th International Conference on Machine
  Learning, {ICML} 2018, Stockholmsm{\"{a}}ssan, Stockholm, Sweden, July 10-15,
  2018}, volume~80 of \emph{Proceedings of Machine Learning Research}, pages
  2309--2318. {PMLR}.

\bibitem[{Kingma and Ba(2015)}]{Kingma2015AdamAM}
Diederik~P. Kingma and Jimmy Ba. 2015.
\newblock Adam: {A} method for stochastic optimization.
\newblock In \emph{3rd International Conference on Learning Representations,
  {ICLR} 2015, San Diego, CA, USA, May 7-9, 2015, Conference Track
  Proceedings}.

\bibitem[{Krogh and Hertz(1992)}]{krogh1992simple}
Anders Krogh and John~A Hertz. 1992.
\newblock A simple weight decay can improve generalization.
\newblock In \emph{Advances in neural information processing systems}, pages
  950--957.

\bibitem[{Lee and Chung(2020)}]{lee2019robust}
Jisoo Lee and Sae-Young Chung. 2020.
\newblock Robust training with ensemble consensus.
\newblock In \emph{International Conference on Learning Representations}.

\bibitem[{Lin et~al.(2016)Lin, Shen, Liu, Luan, and Sun}]{Lin2016NeuralRE}
Yankai Lin, Shiqi Shen, Zhiyuan Liu, Huanbo Luan, and Maosong Sun. 2016.
\newblock Neural relation extraction with selective attention over instances.
\newblock In \emph{Proceedings of the 54th Annual Meeting of the Association
  for Computational Linguistics (Volume 1: Long Papers)}, pages 2124--2133,
  Berlin, Germany. Association for Computational Linguistics.

\bibitem[{Liu et~al.(2017)Liu, Wang, Chang, and Sui}]{Liu2017ASM}
Tianyu Liu, Kexiang Wang, Baobao Chang, and Zhifang Sui. 2017.
\newblock A soft-label method for noise-tolerant distantly supervised relation
  extraction.
\newblock In \emph{Proceedings of the 2017 Conference on Empirical Methods in
  Natural Language Processing}, pages 1790--1795, Copenhagen, Denmark.
  Association for Computational Linguistics.

\bibitem[{Malach and Shalev{-}Shwartz(2017)}]{Malach2017DecouplingT}
Eran Malach and Shai Shalev{-}Shwartz. 2017.
\newblock Decoupling "when to update" from "how to update".
\newblock In \emph{Advances in Neural Information Processing Systems 30: Annual
  Conference on Neural Information Processing Systems 2017, December 4-9, 2017,
  Long Beach, CA, {USA}}, pages 960--970.

\bibitem[{Mayhew et~al.(2019)Mayhew, Chaturvedi, Tsai, and
  Roth}]{mayhew2019named}
Stephen Mayhew, Snigdha Chaturvedi, Chen-Tse Tsai, and Dan Roth. 2019.
\newblock Named entity recognition with partially annotated training data.
\newblock In \emph{Proceedings of the 23rd Conference on Computational Natural
  Language Learning (CoNLL)}, pages 645--655, Hong Kong, China. Association for
  Computational Linguistics.

\bibitem[{Mintz et~al.(2009)Mintz, Bills, Snow, and
  Jurafsky}]{Mintz2009DistantSF}
Mike Mintz, Steven Bills, Rion Snow, and Daniel Jurafsky. 2009.
\newblock Distant supervision for relation extraction without labeled data.
\newblock In \emph{Proceedings of the Joint Conference of the 47th Annual
  Meeting of the {ACL} and the 4th International Joint Conference on Natural
  Language Processing of the {AFNLP}}, pages 1003--1011, Suntec, Singapore.
  Association for Computational Linguistics.

\bibitem[{M{\"u}ller et~al.(2019)M{\"u}ller, Kornblith, and
  Hinton}]{muller2019does}
Rafael M{\"u}ller, Simon Kornblith, and Geoffrey Hinton. 2019.
\newblock When does label smoothing help?
\newblock In \emph{Proceedings of the 33rd International Conference on Neural
  Information Processing Systems}, pages 4694--4703.

\bibitem[{Peng et~al.(2015)Peng, Khashabi, and Roth}]{peng2015solving}
Haoruo Peng, Daniel Khashabi, and Dan Roth. 2015.
\newblock Solving hard coreference problems.
\newblock In \emph{Proceedings of the 2015 Conference of the North American
  Chapter of the Association for Computational Linguistics: Human Language
  Technologies}, pages 809--819.

\bibitem[{Qin et~al.(2018)Qin, Xu, and Wang}]{Qin2018RobustDS}
Pengda Qin, Weiran Xu, and William~Yang Wang. 2018.
\newblock Robust distant supervision relation extraction via deep reinforcement
  learning.
\newblock In \emph{Proceedings of the 56th Annual Meeting of the Association
  for Computational Linguistics (Volume 1: Long Papers)}, pages 2137--2147,
  Melbourne, Australia. Association for Computational Linguistics.

\bibitem[{Ramshaw and Marcus(1995)}]{Ramshaw1995TextCU}
Lance Ramshaw and Mitch Marcus. 1995.
\newblock Text chunking using transformation-based learning.
\newblock In \emph{Third Workshop on Very Large Corpora}.

\bibitem[{Ratner et~al.(2016)Ratner, Sa, Wu, Selsam, and
  R{\'e}}]{Ratner2016DataPC}
Alexander~J. Ratner, Christopher~De Sa, Sen Wu, Daniel Selsam, and C.~R{\'e}.
  2016.
\newblock Data programming: Creating large training sets, quickly.
\newblock \emph{Advances in neural information processing systems},
  29:3567--3575.

\bibitem[{Raykar et~al.(2010)Raykar, Yu, Zhao, Valadez, Florin, Bogoni, and
  Moy}]{raykar2010learning}
Vikas~C Raykar, Shipeng Yu, Linda~H Zhao, Gerardo~Hermosillo Valadez, Charles
  Florin, Luca Bogoni, and Linda Moy. 2010.
\newblock Learning from crowds.
\newblock \emph{Journal of Machine Learning Research}, 11(4).

\bibitem[{Reiss et~al.(2020)Reiss, Xu, Cutler, Muthuraman, and
  Eichenberger}]{reiss2020identifying}
Frederick Reiss, Hong Xu, Bryan Cutler, Karthik Muthuraman, and Zachary
  Eichenberger. 2020.
\newblock Identifying incorrect labels in the {C}o{NLL}-2003 corpus.
\newblock In \emph{Proceedings of the 24th Conference on Computational Natural
  Language Learning}, pages 215--226, Online. Association for Computational
  Linguistics.

\bibitem[{Sang and De~Meulder(2003)}]{Sang2003IntroductionTT}
EF~Tjong~Kim Sang and F~De~Meulder. 2003.
\newblock Introduction to the conll-2003 shared task: Language-independent
  named entity recognition.
\newblock In \emph{Proceedings of CoNLL-2003, Edmonton, Canada}, pages
  142--145. Morgan Kaufman Publishers.

\bibitem[{Shi and Lin(2019)}]{Shi2019SimpleBM}
Peng Shi and Jimmy Lin. 2019.
\newblock Simple bert models for relation extraction and semantic role
  labeling.
\newblock \emph{ArXiv}, abs/1904.05255.

\bibitem[{Song et~al.(2015)Song, Wang, Zhang, Sun, and Yang}]{song2015spectral}
Yangqiu Song, Chenguang Wang, Ming Zhang, Hailong Sun, and Qiang Yang. 2015.
\newblock Spectral label refinement for noisy and missing text labels.
\newblock In \emph{Proceedings of the Twenty-Ninth {AAAI} Conference on
  Artificial Intelligence, January 25-30, 2015, Austin, Texas, {USA}}, pages
  2972--2978. {AAAI} Press.

\bibitem[{Srivastava et~al.(2014)Srivastava, Hinton, Krizhevsky, Sutskever, and
  Salakhutdinov}]{srivastava2014dropout}
Nitish Srivastava, Geoffrey Hinton, Alex Krizhevsky, Ilya Sutskever, and Ruslan
  Salakhutdinov. 2014.
\newblock Dropout: a simple way to prevent neural networks from overfitting.
\newblock \emph{The journal of machine learning research}, 15(1):1929--1958.

\bibitem[{Sukhbaatar et~al.(2015)Sukhbaatar, Bruna, Paluri, Bourdev, and
  Fergus}]{Sukhbaatar2014TrainingCN}
Sainbayar Sukhbaatar, Joan Bruna, Manohar Paluri, Lubomir~D. Bourdev, and
  R.~Fergus. 2015.
\newblock Training convolutional networks with noisy labels.
\newblock In \emph{ICLR}.

\bibitem[{Surdeanu et~al.(2012)Surdeanu, Tibshirani, Nallapati, and
  Manning}]{Surdeanu2012MultiinstanceML}
Mihai Surdeanu, Julie Tibshirani, Ramesh Nallapati, and Christopher~D. Manning.
  2012.
\newblock Multi-instance multi-label learning for relation extraction.
\newblock In \emph{Proceedings of the 2012 Joint Conference on Empirical
  Methods in Natural Language Processing and Computational Natural Language
  Learning}, pages 455--465, Jeju Island, Korea. Association for Computational
  Linguistics.

\bibitem[{Toneva et~al.(2019)Toneva, Sordoni, des Combes, Trischler, Bengio,
  and Gordon}]{Toneva2019AnES}
Mariya Toneva, Alessandro Sordoni, Remi~Tachet des Combes, Adam Trischler,
  Yoshua Bengio, and Geoffrey~J. Gordon. 2019.
\newblock An empirical study of example forgetting during deep neural network
  learning.
\newblock In \emph{7th International Conference on Learning Representations,
  {ICLR} 2019, New Orleans, LA, USA, May 6-9, 2019}. OpenReview.net.

\bibitem[{Wang et~al.(2019{\natexlab{a}})Wang, Liu, Li, Yang, and
  Li}]{Wang2019LearningWN}
Hao Wang, Bing Liu, Chaozhuo Li, Yan Yang, and Tianrui Li. 2019{\natexlab{a}}.
\newblock Learning with noisy labels for sentence-level sentiment
  classification.
\newblock In \emph{Proceedings of the 2019 Conference on Empirical Methods in
  Natural Language Processing and the 9th International Joint Conference on
  Natural Language Processing (EMNLP-IJCNLP)}, pages 6286--6292, Hong Kong,
  China. Association for Computational Linguistics.

\bibitem[{Wang et~al.(2018)Wang, Liu, and Tao}]{Wang2018MulticlassLW}
Ruxin Wang, T.~Liu, and D.~Tao. 2018.
\newblock Multiclass learning with partially corrupted labels.
\newblock \emph{IEEE Transactions on Neural Networks and Learning Systems},
  29:2568--2580.

\bibitem[{Wang et~al.(2017)Wang, Liu, and Tao}]{wang2017multiclass}
Ruxin Wang, Tongliang Liu, and Dacheng Tao. 2017.
\newblock Multiclass learning with partially corrupted labels.
\newblock \emph{IEEE transactions on neural networks and learning systems},
  29(6):2568--2580.

\bibitem[{Wang et~al.(2019{\natexlab{b}})Wang, Ma, Chen, Luo, Yi, and
  Bailey}]{Wang2019SymmetricCE}
Yisen Wang, Xingjun Ma, Zaiyi Chen, Yuan Luo, Jinfeng Yi, and James Bailey.
  2019{\natexlab{b}}.
\newblock Symmetric cross entropy for robust learning with noisy labels.
\newblock In \emph{2019 {IEEE/CVF} International Conference on Computer Vision,
  {ICCV} 2019, Seoul, Korea (South), October 27 - November 2, 2019}, pages
  322--330. {IEEE}.

\bibitem[{Wang et~al.(2020)Wang, Wen, Chen, Huang, Zhang, and
  Zheng}]{Wang2020FindingII}
Zifeng Wang, Rui Wen, Xi~Chen, Shao-Lun Huang, N.~Zhang, and Yefeng Zheng.
  2020.
\newblock Finding influential instances for distantly supervised relation
  extraction.
\newblock \emph{ArXiv}, abs/2009.09841.

\bibitem[{Wang et~al.(2019{\natexlab{c}})Wang, Shang, Liu, Lu, Liu, and
  Han}]{Wang2019CrossWeighTN}
Zihan Wang, Jingbo Shang, Liyuan Liu, Lihao Lu, Jiacheng Liu, and Jiawei Han.
  2019{\natexlab{c}}.
\newblock {C}ross{W}eigh: Training named entity tagger from imperfect
  annotations.
\newblock In \emph{Proceedings of the 2019 Conference on Empirical Methods in
  Natural Language Processing and the 9th International Joint Conference on
  Natural Language Processing (EMNLP-IJCNLP)}, pages 5154--5163, Hong Kong,
  China. Association for Computational Linguistics.

\bibitem[{Wei et~al.(2020)Wei, Feng, Chen, and An}]{Wei2020CombatingNL}
Hongxin Wei, Lei Feng, Xiangyu Chen, and Bo~An. 2020.
\newblock Combating noisy labels by agreement: {A} joint training method with
  co-regularization.
\newblock In \emph{2020 {IEEE/CVF} Conference on Computer Vision and Pattern
  Recognition, {CVPR} 2020, Seattle, WA, USA, June 13-19, 2020}, pages
  13723--13732. {IEEE}.

\bibitem[{Wolf et~al.(2020)Wolf, Debut, Sanh, Chaumond, Delangue, Moi, Cistac,
  Rault, Louf, Funtowicz, Davison, Shleifer, von Platen, Ma, Jernite, Plu, Xu,
  Le~Scao, Gugger, Drame, Lhoest, and Rush}]{wolf2020transformers}
Thomas Wolf, Lysandre Debut, Victor Sanh, Julien Chaumond, Clement Delangue,
  Anthony Moi, Pierric Cistac, Tim Rault, Remi Louf, Morgan Funtowicz, Joe
  Davison, Sam Shleifer, Patrick von Platen, Clara Ma, Yacine Jernite, Julien
  Plu, Canwen Xu, Teven Le~Scao, Sylvain Gugger, Mariama Drame, Quentin Lhoest,
  and Alexander Rush. 2020.
\newblock Transformers: State-of-the-art natural language processing.
\newblock In \emph{Proceedings of the 2020 Conference on Empirical Methods in
  Natural Language Processing: System Demonstrations}, pages 38--45, Online.
  Association for Computational Linguistics.

\bibitem[{Yamada et~al.(2020)Yamada, Asai, Shindo, Takeda, and
  Matsumoto}]{Yamada2020LUKEDC}
Ikuya Yamada, Akari Asai, Hiroyuki Shindo, Hideaki Takeda, and Yuji Matsumoto.
  2020.
\newblock {LUKE}: Deep contextualized entity representations with entity-aware
  self-attention.
\newblock In \emph{Proceedings of the 2020 Conference on Empirical Methods in
  Natural Language Processing (EMNLP)}, pages 6442--6454, Online. Association
  for Computational Linguistics.

\bibitem[{Yang et~al.(2019)Yang, He, Dai, Huang, and
  Chen}]{Yang2019ExploitingND}
Kaijia Yang, Liang He, Xin-yu Dai, Shujian Huang, and Jiajun Chen. 2019.
\newblock Exploiting noisy data in distant supervision relation classification.
\newblock In \emph{Proceedings of the 2019 Conference of the North {A}merican
  Chapter of the Association for Computational Linguistics: Human Language
  Technologies, Volume 1 (Long and Short Papers)}, pages 3216--3225,
  Minneapolis, Minnesota. Association for Computational Linguistics.

\bibitem[{Yu et~al.(2019)Yu, Han, Yao, Niu, Tsang, and Sugiyama}]{Yu2019HowDD}
Xingrui Yu, Bo~Han, Jiangchao Yao, Gang Niu, Ivor~W. Tsang, and Masashi
  Sugiyama. 2019.
\newblock How does disagreement help generalization against label corruption?
\newblock In \emph{Proceedings of the 36th International Conference on Machine
  Learning, {ICML} 2019, 9-15 June 2019, Long Beach, California, {USA}},
  volume~97 of \emph{Proceedings of Machine Learning Research}, pages
  7164--7173. {PMLR}.

\bibitem[{Zeng et~al.(2015)Zeng, Liu, Chen, and Zhao}]{Zeng2015DistantSF}
Daojian Zeng, Kang Liu, Yubo Chen, and Jun Zhao. 2015.
\newblock Distant supervision for relation extraction via piecewise
  convolutional neural networks.
\newblock In \emph{Proceedings of the 2015 Conference on Empirical Methods in
  Natural Language Processing}, pages 1753--1762, Lisbon, Portugal. Association
  for Computational Linguistics.

\bibitem[{Zhang et~al.(2017{\natexlab{a}})Zhang, Bengio, Hardt, Recht, and
  Vinyals}]{Zhang2017UnderstandingDL}
Chiyuan Zhang, Samy Bengio, Moritz Hardt, Benjamin Recht, and Oriol Vinyals.
  2017{\natexlab{a}}.
\newblock Understanding deep learning requires rethinking generalization.
\newblock In \emph{5th International Conference on Learning Representations,
  {ICLR} 2017, Toulon, France, April 24-26, 2017, Conference Track
  Proceedings}. OpenReview.net.

\bibitem[{Zhang et~al.(2018)Zhang, Qi, and Manning}]{Zhang2018GraphCO}
Yuhao Zhang, Peng Qi, and Christopher~D. Manning. 2018.
\newblock Graph convolution over pruned dependency trees improves relation
  extraction.
\newblock In \emph{Proceedings of the 2018 Conference on Empirical Methods in
  Natural Language Processing}, pages 2205--2215, Brussels, Belgium.
  Association for Computational Linguistics.

\bibitem[{Zhang et~al.(2017{\natexlab{b}})Zhang, Zhong, Chen, Angeli, and
  Manning}]{Zhang2017PositionawareAA}
Yuhao Zhang, Victor Zhong, Danqi Chen, Gabor Angeli, and Christopher~D.
  Manning. 2017{\natexlab{b}}.
\newblock Position-aware attention and supervised data improve slot filling.
\newblock In \emph{Proceedings of the 2017 Conference on Empirical Methods in
  Natural Language Processing}, pages 35--45, Copenhagen, Denmark. Association
  for Computational Linguistics.

\bibitem[{Zhang and Sabuncu(2018)}]{Zhang2018GeneralizedCE}
Zhilu Zhang and Mert~R. Sabuncu. 2018.
\newblock Generalized cross entropy loss for training deep neural networks with
  noisy labels.
\newblock In \emph{Advances in Neural Information Processing Systems 31: Annual
  Conference on Neural Information Processing Systems 2018, NeurIPS 2018,
  December 3-8, 2018, Montr{\'{e}}al, Canada}, pages 8792--8802.

\end{thebibliography}
\bibliographystyle{acl_natbib}





\end{document}